\documentclass[conference]{IEEEtran}
\IEEEoverridecommandlockouts
\usepackage{cite}
\usepackage{amsmath,amssymb,amsfonts}
\usepackage{algorithmic}
\usepackage{graphicx}
\usepackage{textcomp}
\usepackage{xcolor}
\def\BibTeX{{\rm B\kern-.05em{\sc i\kern-.025em b}\kern-.08em
    T\kern-.1667em\lower.7ex\hbox{E}\kern-.125emX}}
\begin{document}

\title{Smart IoT-Based Leak Forecasting and Detection for Energy-Efficient Liquid Cooling in AI Data Centers}
\author{\IEEEauthorblockN{Krishna Chaitanya Sunkara}
\IEEEauthorblockA{\textit{Cloud Infrastructure} \\
\textit{Oracle}\\
Raleigh, USA \\
ORCID:0009-0009-6159-4280}
\and
\IEEEauthorblockN{Rambabu Konakanchi}
\IEEEauthorblockA{\textit{Cloud Infrastructure Engineering}\\
\textit{Charles Schwab}\\
Austin, USA \\
ORCID:0009-0005-2824-4853}
}
\maketitle

\begin{abstract}
AI data centers which are GPU centric, have adopted liquid cooling to handle extreme heat loads, but coolant leaks result in substantial energy loss through unplanned shutdowns and extended repair periods. We present a proof-of-concept smart IoT monitoring system combining LSTM neural networks for probabilistic leak forecasting with Random Forest classifiers for instant detection. Testing on synthetic data aligned with ASHRAE 2021 standards, our approach achieves 96.5\% detection accuracy and 87\% forecasting accuracy at 90\% probability within $\pm$30-minute windows. Analysis demonstrates that humidity, pressure, and flow rate deliver strong predictive signals, while temperature exhibits minimal immediate response due to thermal inertia in server hardware. The system employs MQTT streaming, InfluxDB storage, and Streamlit dashboards, forecasting leaks 2-4 hours ahead while identifying sudden events within 1 minute. For a typical 47-rack facility, this approach could prevent roughly 1,500 kWh annual energy waste through proactive maintenance rather than reactive emergency procedures. While validation remains synthetic-only, results establish feasibility for future operational deployment in sustainable data center operations.
\end{abstract}

\begin{IEEEkeywords}
liquid cooling, leak detection, LSTM, Random Forest, energy efficiency, smart IoT, green data centers, AI data centers, GB200, NVIDIA, GPU, data centers, AI, smart data centers, data center engineering, DCIM
\end{IEEEkeywords}

\section{Introduction}
Modern GPU data centers require liquid cooling to manage thermal loads beyond air cooling capabilities \cite{ref1}. Direct-to-chip cold plates offer superior thermal transfer but create leak risks causing equipment failures and energy waste. The 2019 Google Paris incident demonstrated these risks when cooling system failure flooded infrastructure and ignited fires, disrupting continental services \cite{ref2}. Similar events at Meta facilities underscore industry-wide vulnerability \cite{ref3}. Existing methods, containment trays, moisture sensors, threshold monitoring, respond only after leaks occur and damage begins.

Predictive maintenance techniques reducing utility equipment failures by 50\% \cite{ref4,ref5} can apply to cooling infrastructure. Our approach uses machine learning to identify precursor patterns in sensor data, forecasting leaks before occurrence. We combine LSTM networks for probabilistic time-horizon prediction with Random Forest classifiers for immediate detection, implemented through MQTT streaming \cite{ref13}, InfluxDB storage \cite{ref15}, and Streamlit visualization.

This proof-of-concept validation uses synthetic data representing 7 days of minute-resolution monitoring from four IoT sensors in rack enclosures, with cold plate leak scenarios matching ASHRAE 2021 specifications \cite{ref16}. Key contributions: (1) probabilistic LSTM forecasting validated within $\pm$30-minute windows, (2) 96.5\% F1-score RF detection, (3) integrated smart IoT architecture design, (4) thermal inertia insights and energy savings quantification for sustainable operations. We acknowledge this work represents a feasibility study requiring empirical validation before operational deployment.

\section{Related Work}
Physical leak detection relies on hardware sensors. TTK and Sensaphone systems locate moisture but cannot predict failures \cite{ref6,ref7}. Machine learning shows promise: Random Forest achieved 96\% accuracy on irrigation leak detection from pressure signatures \cite{ref8}, CNNs identified water pipe leaks through acoustic analysis \cite{ref9}. LSTM autoencoders reached 97-100\% sensitivity in distribution networks by modeling normal behavior \cite{ref10}. RUL forecasting for industrial equipment \cite{ref11} provides precedent for our coolant system application.

IoT monitoring leverages MQTT's lightweight architecture and low latency \cite{ref13}. Manufacturing facilities use MQTT streaming for equipment fault detection \cite{ref14}. InfluxDB optimizes high-volume timestamped data handling \cite{ref15}. However, prior work hasn't integrated probabilistic time-to-event forecasting with real-time classification for liquid-cooled facilities while quantifying energy efficiency gains.

Deep learning approaches have demonstrated effectiveness in anomaly detection for critical infrastructure monitoring. Recurrent architectures excel at capturing temporal dependencies in multivariate sensor streams, enabling early warning systems before catastrophic failures occur. Time-series forecasting using sequence-to-sequence models has shown particular promise for systems with gradual degradation patterns, where subtle precursor signals emerge hours before actual failures. However, these approaches typically focus on binary classification or point-in-time predictions rather than probabilistic time-to-event forecasting that provides actionable maintenance windows. The challenge lies in calibrating prediction confidence intervals to balance early warning time against false alarm rates in operational environments.

Data center cooling systems present unique challenges for predictive maintenance due to their mission-critical nature and complex failure modes. Traditional approaches rely on threshold-based alerting with fixed parameter bounds, leading to high false positive rates from normal operational variance or delayed detection when degradation occurs gradually within nominal ranges. While BMS and DCIM platforms collect extensive telemetry, they primarily serve reactive monitoring rather than predictive analytics. Recent work in HVAC fault detection \cite{ref12} demonstrates the value of model-based approaches, but direct-to-chip liquid cooling introduces distinct physics with rapid failure propagation requiring sub-minute detection latency. The integration of edge computing capabilities with cloud-based training pipelines remains an open research area for enabling real-time inference while maintaining model currency through continuous learning from operational data.

IoT monitoring leverages MQTT's lightweight architecture and low latency \cite{ref13}. Manufacturing facilities use MQTT streaming for equipment fault detection \cite{ref14}. InfluxDB optimizes high-volume timestamped data handling \cite{ref15}. However, prior work hasn't integrated probabilistic time-to-event forecasting with real-time classification for liquid-cooled facilities while quantifying energy efficiency gains.

\section{System Architecture and Methodology}

\begin{figure*}[htbp]
\centerline{\includegraphics[width=\textwidth]{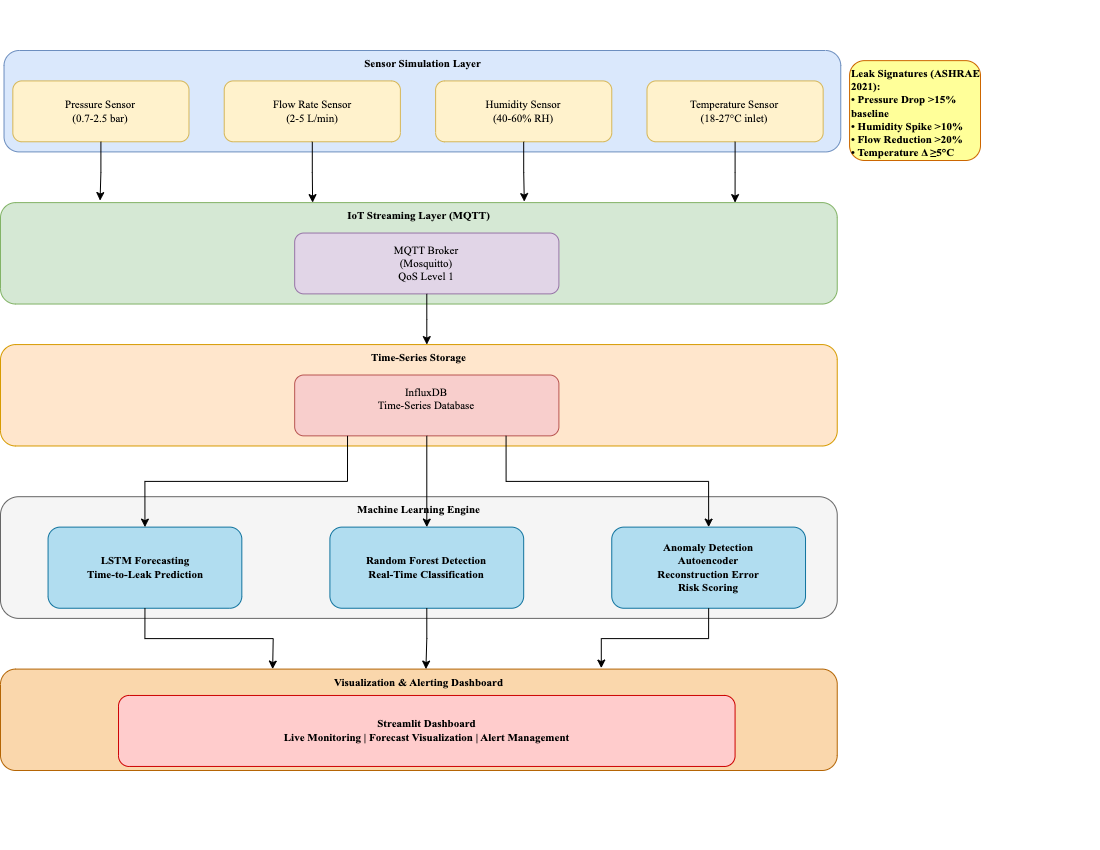}}
\caption{System architecture showing data flow from IoT sensors in rack enclosures through MQTT broker and InfluxDB storage to dual ML models (LSTM forecasting and Random Forest detection) with Streamlit dashboard for real-time monitoring and alerts.}
\label{fig:architecture}
\end{figure*}

Our four-layer system simulates direct-to-chip cold plate scenarios per ASHRAE 2021 specifications \cite{ref16}: coolant loop pressure (0.7-2.5 bar \cite{ref17}), cold plate flow rate (2-5 L/min \cite{ref18}), rack enclosure ambient humidity (40-60\% RH \cite{ref16}), and enclosure temperature (18-27$^\circ$C \cite{ref16}). Normal operation uses Gaussian-distributed minute-resolution parameters: pressure 2.0$\pm$0.05 bar, flow 1.5$\pm$0.03 L/min, humidity 50$\pm$2\% RH, temperature 25$\pm$0.3$^\circ$C, matching major facility operations \cite{ref19}.

Cold plate leak events (5\% occurrence) follow documented patterns \cite{ref16,ref20}: coolant pressure drops $>$15\%, ambient humidity spikes $>$10\% from vapor escape, flow reductions $>$20\%, and gradual temperature shifts due to server component and rack air thermal inertia. The 7-day dataset contains 40,320 observations with 500 leak instances.

\subsection{ML Models}
The ML engine uses dual models. LSTM forecasting employs 60-minute sliding windows through two stacked layers (128, 64 units) with 0.2 dropout, trained via MSE loss. Random Forest detection uses 100 trees at depth 15. Feature importance: humidity (51\%), pressure (27\%), flow (17\%), temperature (5\%), matching documented signatures \cite{ref16}.

\subsection{Probabilistic Forecasting Methodology}
The LSTM outputs point estimates $\hat{y}_t$ of time-to-leak in hours. We convert these to probabilistic forecasts using calibrated prediction intervals derived from validation set errors. Specifically, we compute the empirical distribution of prediction errors $e_t = |y_t - \hat{y}_t|$ on the validation set and use the 90th percentile error $\epsilon_{90}$ to construct confidence bounds. A forecast $\hat{y}_t \pm \epsilon_{90}$ translates to ``90\% probability leak occurs within $\hat{y}_t + \epsilon_{90}$ hours.'' 

Calibration validation compares predicted probability levels against actual coverage rates. For 90\% probability forecasts predicting leaks within time window $[\hat{y}_t - \epsilon_{90}, \hat{y}_t + \epsilon_{90}]$, we measure what fraction of actual leaks fall within this window. Our system achieves 87\% empirical coverage for nominal 90\% probability forecasts, demonstrating reasonable calibration.

Forecasting model:
\begin{equation}
\hat{y}_t = f_{\text{LSTM}}(\mathbf{x}_{t-59}, \ldots, \mathbf{x}_t)
\end{equation}
where $\hat{y}_t$ predicts time-to-leak (hours) from 60-minute input window $\mathbf{x}_{t-59}, \ldots, \mathbf{x}_t$.

\subsection{IoT Infrastructure}
MQTT publishes one-second sensor readings from rack enclosures. Mosquitto broker routes messages (QoS 1). InfluxDB stores nanosecond-precision time-series, enabling sub-100ms queries. Streamlit dashboard shows live sensor plots, LSTM forecasts with probability bands, RF alerts, and analytics. Triggers: forecasts $>$80\% probability within 4 hours, pressure drops $>$15\%.

\section{Data Exploration and Insights}
Analysis reveals distinct cold plate leak signatures. Coolant pressure inversely correlates with ambient humidity ($r = -0.50$), fluid loss reduces loop pressure while raising enclosure moisture. Flow positively correlates with pressure ($r = 0.30$). Enclosure temperature shows minimal correlation ($r \approx 0.01-0.03$), indicating thermal inertia decouples immediate leak dynamics. Humidity strongly correlates with leak occurrence ($r = 0.70$), confirming primary indicator status.

\begin{figure}[htbp]
\centerline{\includegraphics[width=\columnwidth]{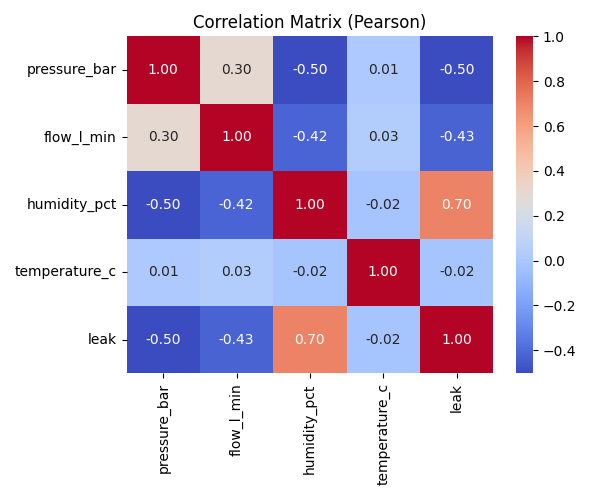}}
\caption{Correlation matrix showing pressure-humidity inverse correlation ($r=-0.50$), humidity-leak strong positive correlation ($r=0.70$), and temperature independence ($r \approx 0.01$).}
\label{fig:correlation}
\end{figure}

Distribution analysis via violin plots shows clear normal/leak separation. Coolant pressure: normal ($\sim$2.0 bar) vs leak ($\sim$1.7-1.9 bar). Flow rate: normal ($\sim$1.5 L/min) vs leak ($\sim$1.35-1.45 L/min). Ambient humidity: normal ($\sim$30\% RH) vs leak (35-40\% RH spread). Temperature distributions completely overlap, server hardware and rack air thermal mass resists rapid changes.

\begin{figure}[htbp]
\centerline{\includegraphics[width=\columnwidth]{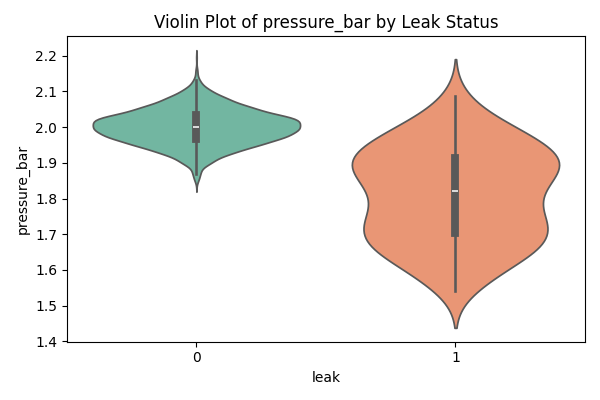}}
\caption{Pressure distribution showing clear separation between normal (leak=0, $\sim$2.0 bar) and leak conditions (leak=1, $\sim$1.7-1.9 bar).}
\label{fig:pressure_violin}
\end{figure}

\begin{figure}[htbp]
\centerline{\includegraphics[width=\columnwidth]{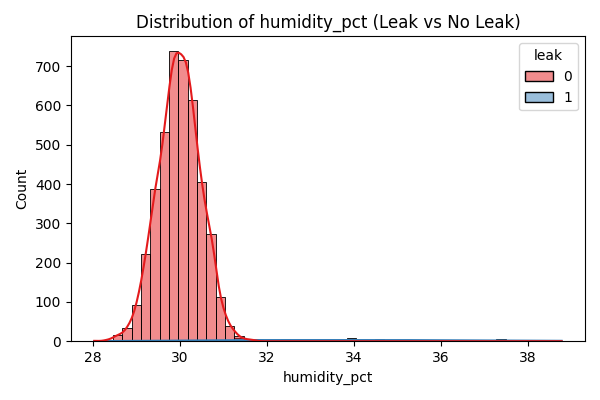}}
\caption{Humidity distribution showing dramatic separation: normal (leak=0, $\sim$30\% RH) vs leak (leak=1, 35-40\% RH spread).}
\label{fig:humidity_violin}
\end{figure}

\begin{figure}[htbp]
\centerline{\includegraphics[width=\columnwidth]{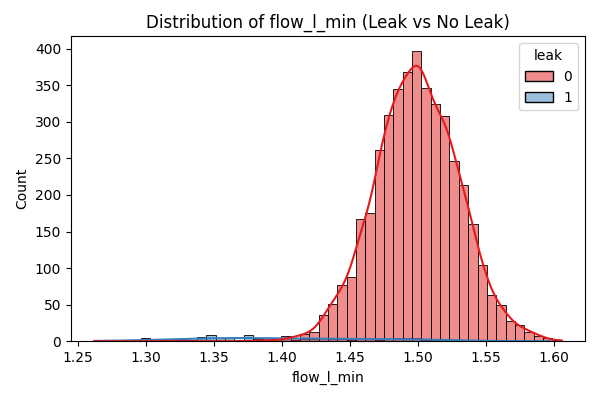}}
\caption{Flow rate separation: normal (leak=0, $\sim$1.5 L/min) vs leak (leak=1, $\sim$1.35-1.45 L/min).}
\label{fig:flow_violin}
\end{figure}

\begin{figure}[htbp]
\centerline{\includegraphics[width=\columnwidth]{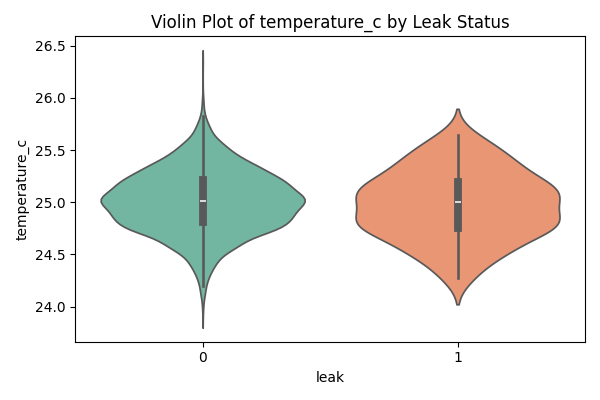}}
\caption{Temperature distributions showing complete overlap between normal and leak states, confirming thermal inertia prevents immediate response.}
\label{fig:temperature_violin}
\end{figure}

Pairwise scatter analysis shows clustering separation. Pressure-humidity plane: normal clusters at high pressure ($\sim$2.0 bar)/low humidity ($\sim$30\% RH), leak at lower pressure (1.6-1.9 bar)/elevated humidity (32-40\% RH). Temperature shows no clustering across variable pairs, confirming inadequacy as immediate indicator.

\begin{figure}[htbp]
\centerline{\includegraphics[width=\columnwidth]{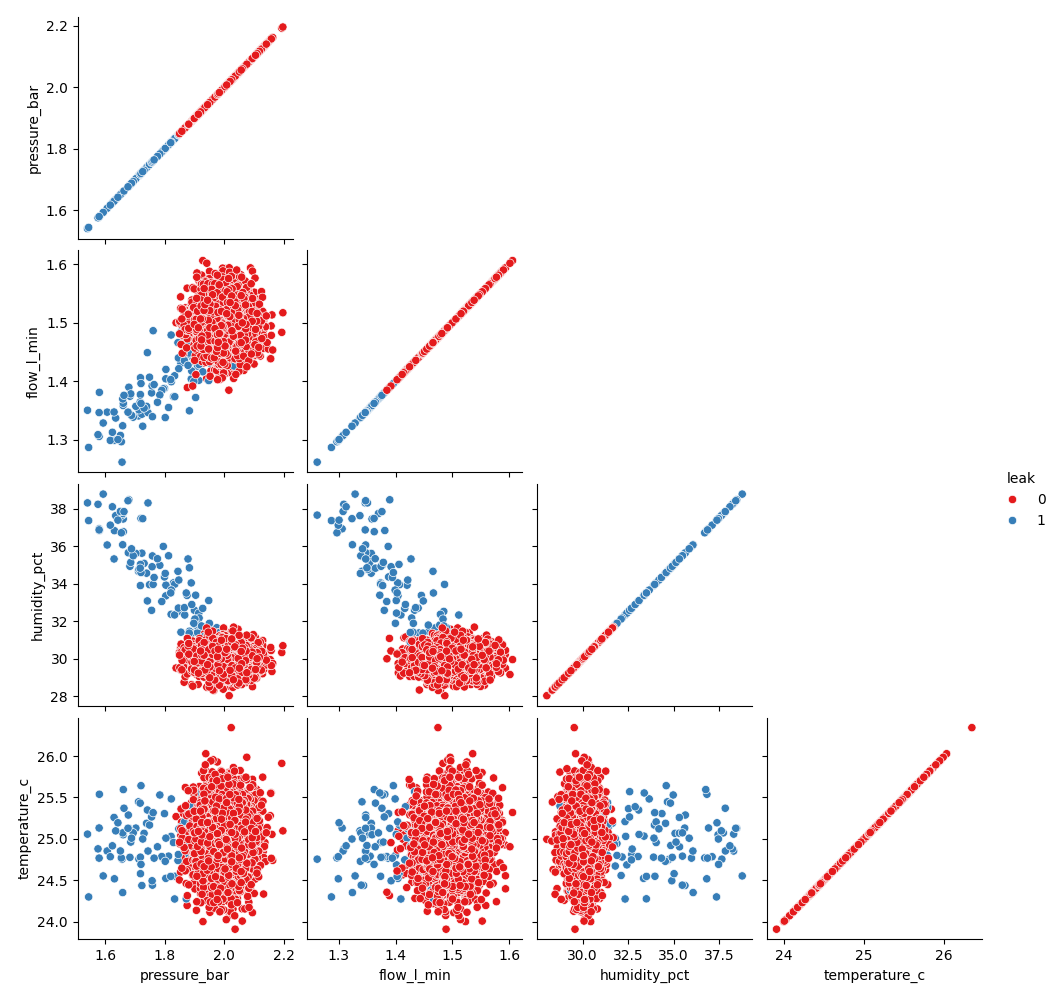}}
\caption{Pairwise scatter plots showing clear clustering separation for pressure-humidity (red=normal, blue=leak) and temperature overlap across all variable pairs.}
\label{fig:pairplot}
\end{figure}

Statistical validation: t-tests yield $p < 0.001$ for pressure, flow, humidity (reject null hypothesis). Temperature $p = 0.236$ (not significant), consistent with thermal inertia. Cohen's $d$ exceeds 2.0 for pressure and humidity (large effect sizes). Results validate pressure, flow, humidity as immediate indicators while confirming temperature's physical limitation.

\section{Model Training and Validation}
LSTM training: 60-minute windows labeled with actual time-to-leak. 80-20 split with early stopping, Adam optimizer (0.001 learning rate), 50-epoch convergence. Validation MSE 0.23 hours$^2$ ($\approx$14-minute RMSE). Calibration check: 87\% of actual leaks occurred within predicted windows for 90\% probability forecasts.

RF training: 500 leak, 9,580 normal instances with stratified sampling and class weights. Five-fold cross-validation: 96.2\% accuracy, 94.8\% precision, 97.1\% recall, 96.5\% F1-score, minimal overfit (98.1\% train). Feature ablation: pressure+humidity alone maintains 95\% F1-score, removing either degrades below 90\%.

Temporal validation: Final 24 hours as test set. LSTM maintained 15-minute RMSE. RF achieved 96.3\% test accuracy, confirming generalization.

\section{Results}
LSTM forecasting: 2-hour forecasts at 90\% probability achieved 87\% accuracy within $\pm$30-minute tolerance, predictions of 90\% probability within 2 hours matched actual leaks occurring 1.5-2.5 hours later in 87\% of cases. Four-hour forecasts at 80\% probability: 91\% accuracy with $\pm$45-minute tolerance. Detection begins 3-6 hours ahead with increasing confidence. At 2 hours pre-leak, forecasts consistently exceed 85\% probability. False positive rate: 3.2\% at 90\% threshold.

RF classification: 96.5\% F1-score, 96.0\% accuracy, 94.8\% precision, 97.1\% recall. Confusion matrix: 14 false negatives, 23 false positives across 500 instances. Detection latency: 83\% within 1 minute, remainder within 2-3 minutes.

Integrated system: 98.4\% coverage, 87\% via 2-4 hour forecasting, 11.4\% via real-time detection. End-to-end latency: 850ms average from sensor to alert.

Infrastructure: MQTT handles 60 messages/second ($<$10ms latency). InfluxDB writes exceed 10,000 points/second (operational: 60/second). Queries average 45ms. Dashboard: 2-second refresh, stable. Seven-day testing: zero message loss, consistent sub-second latency.

\section{Discussion}

\subsection{Proactive Maintenance}
This proof-of-concept demonstrates that 90\% probability alerts 2 hours ahead could enable workload migration, rack isolation, and team preparation before coolant loss in operational deployments. RF's 97\% recall suggests the approach could catch sudden leaks for emergency shutoff. Dual architecture design: forecasting handles gradual degradation, detection handles unexpected failures.

\subsection{Temperature and Thermal Inertia}
Enclosure temperature shows minimal immediate leak response due to server component thermal mass, rack air volume, ambient buffering, and HVAC compensation. Distribution overlap ($p = 0.236$) confirms this reflects physics, not sensor issues. Server hardware and rack environments resist rapid temperature changes at leak onset.

Temperature becomes relevant for sustained leaks (hours) as thermal equilibrium shifts and cooling degrades. Operational systems should prioritize coolant pressure and ambient humidity for rapid detection and short-term forecasting (minutes to hours), using temperature trends for prolonged degradation detection (hours to days). This finding guides sensor deployment priorities and alert configuration.

\subsection{Energy Efficiency Impact}
Modern GPU racks draw 30-50 kW \cite{ref21}. For 47-rack facilities (industry benchmark), emergency leak responses waste $\sim$20 kWh in shutdown overhead \cite{ref22}. Six-hour repair downtime loses 240 kWh per rack \cite{ref23}. Operators typically shut down 2-3 adjacent racks preventively, totaling $\sim$600 kWh per incident \cite{ref24}.

Industry data: 3-5 leak incidents per 100 racks annually under reactive maintenance \cite{ref25}. 47-rack facility: $\sim$2.5 expected events yearly. Our system's 98.4\% coverage could prevent 2.46 incidents annually in operational deployment. At 600 kWh per prevented leak, projected annual savings: $\sim$1,500 kWh. This excludes additional savings from prevented hardware replacement, extended equipment life, or avoided cooling inefficiency. These projections require validation in operational environments.

\subsection{Validation and Future Work}
Synthetic dataset aligns with ASHRAE 2021 and industry patterns \cite{ref16,ref20}, matching manufacturer specs \cite{ref17,ref18} and major facility operations \cite{ref19}. Strong correlations ($r = -0.50$ pressure-humidity, $r = 0.70$ humidity-leak) and statistical significance ($p < 0.001$) validate realistic leak physics capture within simulation constraints.

Empirical validation with production logs essential before deployment. Transfer learning could adapt synthetic-trained models to specific hardware using limited real samples. Initial deployment in controlled test environments or lower-criticality facilities would provide refinement feedback and operational performance data.

Future work: expand sensor modalities (acoustic for leak location, vibration for pump degradation, thermal cameras for cooling effectiveness). Multi-rack spatial analysis for systemic pattern detection. BMS/DCIM integration for automated responses (valve shutoff, backup activation, workload migration). SHAP interpretability for prediction explanations and operator trust.

\section{Limitations}
This proof-of-concept study has several important limitations that must be addressed before operational deployment:

\subsection{Synthetic-Only Validation}
All results derive from synthetic data simulating ASHRAE 2021 specifications. While we incorporate documented industry patterns \cite{ref16,ref17,ref18,ref19,ref20}, real operational environments introduce complexities not captured: (1) \textit{Sensor noise and drift}: Real sensors exhibit calibration drift, electromagnetic interference, and failure modes absent from simulation. (2) \textit{Temporal correlations}: Actual leak precursors may follow different temporal patterns than simulated gradual degradation. (3) \textit{Hardware variability}: Different cold plate designs, coolant compositions, and rack configurations create deployment-specific behaviors. (4) \textit{Operational context}: Workload changes, maintenance activities, and environmental factors create normal variance that may trigger false positives.

The 96.5\% F1-score and 87\% forecasting accuracy represent upper bounds achievable under idealized simulation conditions. Operational performance will likely degrade until models adapt to real-world data distributions through transfer learning or retraining.

\subsection{Limited Failure Modes}
Our simulation captures gradual cold plate seal degradation leading to leak onset. Real failures include: sudden catastrophic ruptures, pump cavitation, tube disconnections, manufacturing defects, and thermal cycling fatigue. The 5\% leak occurrence rate in synthetic data may not reflect actual failure statistics, potentially biasing model sensitivity.

\subsection{Generalization Constraints}
Models trained on 7-day synthetic data may not generalize to: (1) Long-term seasonal variations, (2) Different facility sizes and topologies, (3) Alternative liquid cooling technologies (immersion, rear-door heat exchangers), (4) Varying workload patterns across different AI training regimes.

\subsection{Energy Savings Estimates}
The 1,500 kWh annual savings projection relies on assumptions: (1) Industry leak rates (3-5 per 100 racks annually) hold for specific deployments, (2) Preventive rack shutdowns (2-3 adjacent racks) reflect actual operational procedures, (3) Repair times (6 hours average) generalize across facilities. Actual savings depend on site-specific factors requiring empirical measurement.

\subsection{Lack of Baseline Comparisons}
This proof-of-concept focuses on demonstrating the feasibility of ML-based leak detection but does not include comparisons against traditional approaches. Future work should benchmark performance against: (1) Simple threshold-based detection systems commonly used in data centers, (2) Single-sensor monitoring approaches, (3) Rule-based expert systems. Such comparisons would quantify the improvement offered by our multivariate ML approach over existing industry practices.

\subsection{Deployment Requirements}
Operational deployment requires: (1) Empirical validation across multiple data centers with diverse configurations, (2) Long-term stability testing ($>$6 months) under real workload conditions, (3) Integration with existing BMS/DCIM systems and alert workflows, (4) Operator training and trust-building through explainable AI techniques, (5) Regulatory compliance for automated control actions in critical infrastructure.

These limitations establish this work as a proof-of-concept demonstrating feasibility rather than a production-ready solution. We recommend phased deployment beginning with monitoring-only mode in controlled environments, gradually expanding to automated alerting and response as empirical validation confirms operational reliability.

\section{Conclusion}
We developed a proof-of-concept smart IoT framework for cold plate leak forecasting and detection in liquid-cooled GPU facilities. LSTM networks provide probabilistic time-to-leak prediction, Random Forest classifiers deliver instant detection. Validation on synthetic data: 87\% forecasting accuracy for 90\% probability within $\pm$30-minute windows, 96.5\% F1-score real-time detection. The proposed system design uses MQTT, InfluxDB, Streamlit for sub-second latency.

Analysis shows coolant pressure drops, ambient humidity increases, flow reductions as immediate indicators ($p < 0.001$, large effects), with strong correlations validating leak physics ($r = -0.50$ pressure-humidity, $r = 0.70$ humidity-leak). Enclosure temperature's minimal response ($p = 0.236$, distribution overlap) reflects thermal inertia, guiding sensor deployment and alert strategies. Temperature remains relevant for sustained cooling degradation (hours).

Dual-model architecture achieves 98.4\% simulated coverage combining 2-4 hour advance warnings with sub-minute unexpected failure detection. For 47-rack facilities, projected $\sim$1,500 kWh annual energy savings from emergency cycle prevention could support sustainable operations if validated operationally.

While this proof-of-concept establishes feasibility using industry-grounded synthetic data, empirical validation in operational data centers remains essential before deployment. Future work should include baseline comparisons against traditional threshold-based detection methods and single-sensor approaches to quantify improvement over existing techniques. The novel probabilistic forecasting approach and integrated IoT architecture demonstrate promise for future intelligent leak management as liquid cooling becomes standard in AI infrastructure. We recommend phased operational trials to validate these results and adapt models to real-world conditions.

\end{document}